\documentclass{article}
\usepackage{spconf,amsmath,graphicx}
\usepackage{subfig}
\usepackage{courier}
\usepackage{pifont}
\usepackage{graphicx}
\usepackage{hyperref}
\usepackage{amssymb}
\usepackage{amsthm} 
\usepackage{amsmath}
\usepackage{epsfig}
\usepackage{epstopdf}
\usepackage{algorithm}
\usepackage{algorithmic}
\usepackage{array,booktabs}
\newcolumntype{M}[1]{>{\centering\arraybackslash}m{#1}}
\usepackage{multirow} 
\usepackage{mathtools}

\newcommand{\cmark}{\ding{51}}%
\newcommand{\xmark}{\ding{55}}%
\newcommand{\real}{\mathbb{R}}

\title{Diversity Promoting Online Sampling for Streaming Video Summarization}
\name{Rushil Anirudh$^{1}$, Ahnaf Masroor$^3$, Pavan Turaga$^{2,3}$	}
\address{$^1$IBM Almaden Research Center, San Jose, CA. \\ $^2$School of Arts, Media, \& Engineering,
$^3$School of Electrical, Computer, \& Energy Engineering,\\
Arizona State University.}

\begin{document}
%
\maketitle
\begin{abstract}
Many applications benefit from sampling algorithms where a small number of well chosen samples are used to generalize different properties of a large dataset. In this paper, we use diverse sampling for streaming video summarization. Several emerging applications support streaming video, but existing summarization algorithms need access to the entire video which requires a lot of memory and computational power. We propose a memory efficient and computationally fast,  online algorithm that uses competitive learning for diverse sampling. Our algorithm is a generalization of online K-means such that the cost function reduces clustering error, while also ensuring a diverse set of samples. The diversity is measured as the volume of a convex hull around the samples. Finally, the performance of the proposed algorithm is measured against human users for 50 videos in the VSUMM dataset. The algorithm performs better than batch mode summarization, while requiring significantly lower memory and computational requirements.

\end{abstract}
\begin{keywords}
Video Summarization, Online Algorithms, Sampling, Streaming Video
\end{keywords}
\section{Introduction}
\label{sec:intro}
\begin{figure}[!tb]
\centering
\includegraphics[trim={50 100 50 90},clip,width = 3.5in]{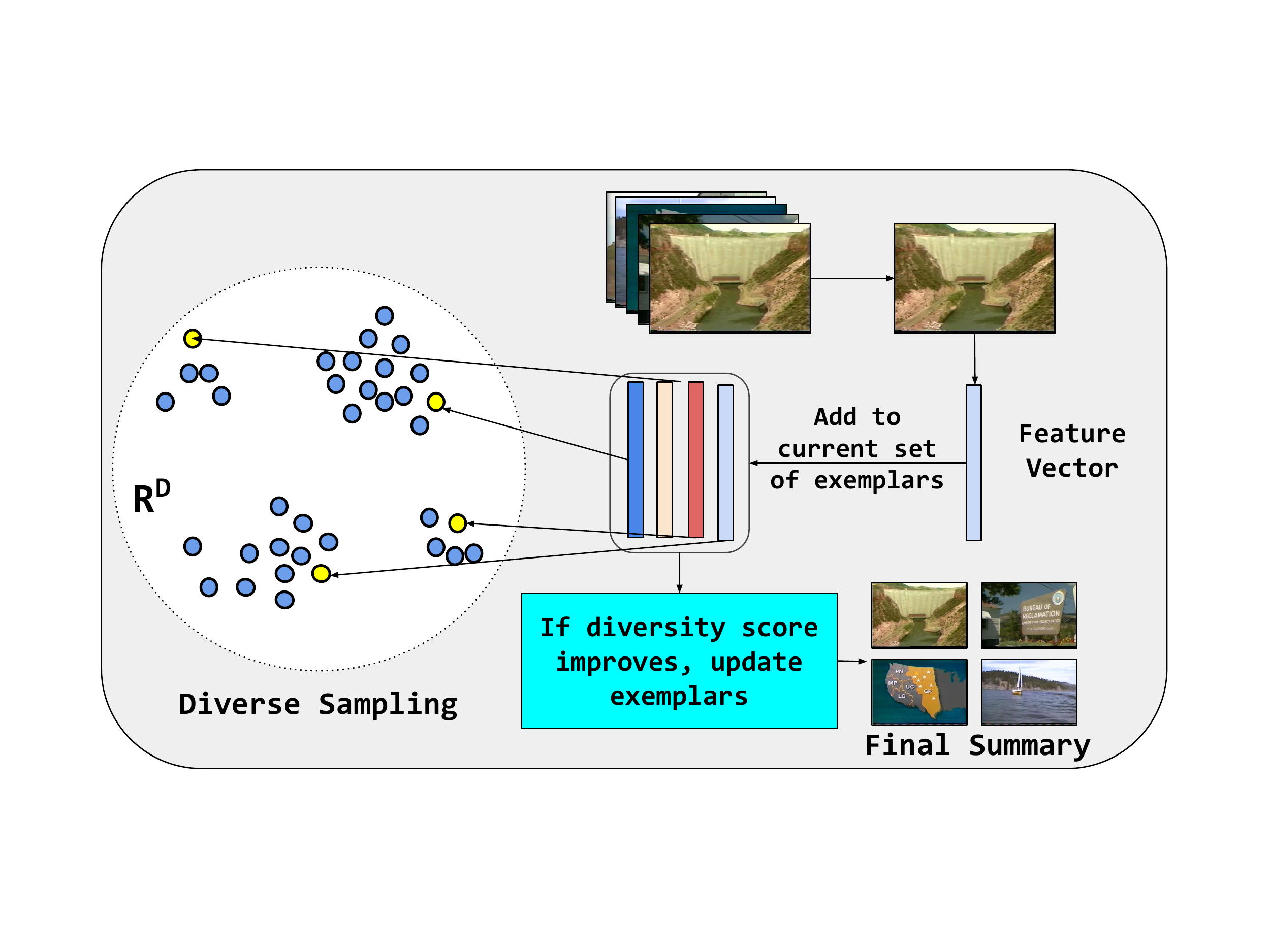}
\caption{\small{\bf Overview of our system} for online video summarization.}
\vspace{-20pt}
\label{fig:overview}
\end{figure}

Smart sampling algorithms are useful in applications where computational or memory resources are limited. In such scenarios, a small number of well chosen samples can be used to generalize properties of an entire dataset for training \cite{settles2012active}, labeling \cite{AnirudhWACV2014}, or other learning problems \cite{ShroffTC10,Yan2015}. We are interested in video summarization, which can be broadly defined as  the problem of picking the $K$ best frames/shots/segments of a video. The challenge in summarizing a video is to define an appropriate cost function, since it can be very subjective based on the application. Almost all video summarization algorithms today work after the fact, i.e. they assume access to the entire video at a time. However, there are many emerging applications with high definition streaming video, where there is a need to perform summarization with little or no memory overhead such as videos on mobile platforms etc. In this work we propose a online generalization of the video summarization problem so that it can work while accessing a single frame at a time, as shown in figure \ref{fig:overview}.  We formulate summarization as a diverse sampling problem, which picks the most \emph{diverse} set of samples from a dataset. This approach is inspired by Video Precis \cite{ShroffTC10}, a batch-mode algorithm, that modifies the $K$-means clustering cost to include the \emph{diversity} of centers in addition to the standard $\ell_2$ clustering error. The additional diversity term improves sampling by making the algorithm less sensitive to large and dense clusters, unlike K-means. In the context of summarization, this results in a summary that samples from all key events. An effective video summarization algorithm trades-off between representing most of the video and picking unique and/or interesting frames that may occur sparsely. Our algorithm has memory requirements in the order of $\mathcal{O}(K)$, where $K$ is the length of the summary, typically in the range of 10-100. This is much better than existing approaches, which require \emph{at least} $\mathcal{O}(N)$, the computational complexity is also linear in $N$, compared to quadratic complexity for comparable approaches. 
\let\thefootnote\relax\footnotetext{The work of A.M was supported by REU supplement to NSF CAREER grant 1452163 and the work of R.A and P.T was supported by NSF grants 1452163 and 1320267.}
\noindent Existing approaches for batch-mode summarization have used different strategies to define importance scores for events in a video. For example, the work in \cite{ghosh2012discovering} focuses on ego-centric video and uses visual cues that humans often use such as the position of the object within the frame. As a result, any object in the focus of the user is given high importance. The idea of important objects from a single view point, has also been generalized for generic videos \cite{khosla2013large}. In many videos, there is a lot of content in video transitions, which can be omitted using priors learned from the web \cite{khosla2013large}. Adaptive or dynamic video summarization does not enforce a fixed summary length and adapts the length of the summary based on the information within the video \cite{ChakrabortyWACV2015}. Online summarization for videos has remained largely unexplored -- the work in \cite{almeida2012vison} proposes to use a user-customizable summarization which allows the user to specify quality of the summary and also the time available for the process. This technique enhances the user experience and speeds up the process by creating the summary as an online task, saving time. In contrast, we propose an online algorithm that can work with any kind of image/video features, while having access to a single frame at a time. We propose a generalization to the online K-means clustering algorithm, that also includes a \emph{diversity  bias}. This ensures that each sample is assigned to a center that is close to it while also satisfying the diversity constraint. In a special case, our algorithm reduces to the online K-medoids clustering algorithm. We show that the proposed algorithm is able to summarize videos significantly better than several comparable baselines, at significantly lesser computational cost. We show extensive evaluation on a dataset of $50$ videos \cite{VSUMMweb,Avila2011} and perform a comparison with human-user generated summaries.

\section{Problem Formulation}
\label{sec:formulation}
The summarization problem can be stated as follows: given a set of frames from a video $X = \{x_1,x_2,\dots,x_n\}, x_i\in \real^D$, pick the most \emph{representative} $K$ points, $\boldsymbol{\mu} = \{\mu_1,\mu_2,\dots,\mu_K\}$ from the set. We will refer to these representatives as exemplars. The $x_i$'s can be a feature or a set of features extracted per frame from the video. Summarization or diverse sampling is similar to clustering in many ways, and the clustering analogy is useful to illustrate our algorithm. For example, $K$-means (or $K$ medoids) is a sampling algorithm when the centers are the samples, chosen by minimizing the $\ell_2$ clustering error. In online $K$-means, a \emph{competition} is held between centers to determine who `wins' the current sample, determined by which center is the closest to the current sample in the sense of the Euclidean norm. The winning center is moved in the direction of the sample, by a small amount governed by the learning rate, $\alpha \in [0,1]$. That is, for a winning center $\mu_k$ and the $i^{th}$ point $x_i$, the updated center is given by $\widehat{\mu}_k = \mu_k + \alpha~(x_i-\mu_k).$ 

However K-means can be very biased towards larger clusters, leading to poor summaries. To overcome this, we modify the clustering error term to include a notion of diversity bias which forces the centers apart, instead of having several centers in a single large cluster. The diversity bias is similar to the conscience bias \cite{Desieno1988conscience} that can be used to generate equi-probable clusters, where the bias discourages a center from winning too often. Instead, the diversity bias promotes updating centers that improve the overall diversity. The modified cost function resembles  the one used in Video Precis \cite{ShroffTC10} for batch-mode summarization.  In our algorithm the criterion to determine the winning center for the $i^{th}$ round is given by $ \hat{k} = \operatorname*{arg\,min}_k d(k)$, where $d(k)$ is given by:
\begin{equation}
\label{eq:distfunc}
\small{d(k) = \beta~||x_i-\mu_k||^2+ C(1-\beta)~\texttt{divscore}(\boldsymbol{\mu}_{k\leftarrow i})-\zeta},
\end{equation}
where $(\boldsymbol{\mu}_{k\leftarrow i})$ denotes the set of centers, when the $k^{th}$ center is replaced by the current data point $x_i$, $C$ is a normalizing factor that ensures all data points are given the same importance, and $\zeta$ is the previous maximum diversity score computed using the function -- \texttt{divscore}(~).

\subsection{Diversity Measure} 
\label{sec:divmeasure}
The choice of the function \texttt{divscore}(~), in equation \eqref{eq:distfunc} is important since it significantly influences the final summary. Diversity can be measured using dispersion measures such as the sample variance of the centers, as in \cite{ShroffTC10}. However, we observed that it can encourage a grouping behaviour, where a pair of centers is close to each other but far away from the rest of the centers. 

\noindent \textbf{Volume of the convex hull:} We propose to use the volume of the convex polytope formed by the centroids, as our diversity score. A convex  polytope $P$ is the convex hull $conv(\boldsymbol{\mu})$ for a finite set of centers. Computing the volume is hard in general and computationally expensive \cite{bueler2000exact}, especially when the points are in higher dimensions \cite{Barber1996quickhull}. Fortunately in lower dimensions its time efficient, and there are several standard implementations. We use the \texttt{qHull,convexhulln} functions in MATLAB \cite{Barber1996quickhull}. For high dimensional features, we map the centers to $\real^d, d << D.$ and then compute the volume of the convex-hull in $\real^d$. Although this may not reflect the true volume, it is an approximation that works well in practice.
\begin{algorithm}[!htb]
\caption{Online Diverse Sampling}
\label{algo:oprecis}
\begin{algorithmic}[1]
\STATE {\bf Input}: \small{Currrent frame $x_i \in \real^D$, Number of exemplars $K$}
\STATE {\bf Output}: Exemplars $\boldsymbol{\mu} = \{\mu_1,\mu_2,\dots,\mu_K\}$ 
\IF{$i<K$}
\STATE $\boldsymbol{\mu}(i) = x_i$ // Initialization
\STATE exemplar\_idx$(i) = i$
\ELSE
\STATE $C = 10~i$ //normalizing factor
\STATE $ \zeta$ = $\texttt{divscore}(\boldsymbol{\mu}$)) (see sec \ref{sec:divmeasure})
\FOR{$k\leftarrow[1\dots K]$}
\STATE $div(k) = \texttt{divscore}(\boldsymbol{\mu}_{k\leftarrow i}) - \zeta$ 
\STATE d$(k) = \beta~\lVert x_i-\mu_k \rVert_2 - C(1-\beta)~(div(k)$)
\ENDFOR
\STATE $ j = \operatorname*{arg\,min}_k d(k) $
\IF{$ div(j)> \zeta$}
\STATE exemplar\_idx($j$) $= i,\boldsymbol{\mu}(j) = x_i$ //update
\STATE $\zeta = div(idx)$
\STATE divcost($i$) = $\zeta$
\ENDIF
\ENDIF
\end{algorithmic}
\end{algorithm}

\begin{figure}
\centering
\includegraphics[clip = true,trim=40mm 10mm 40mm 15mm,width=0.90\linewidth]{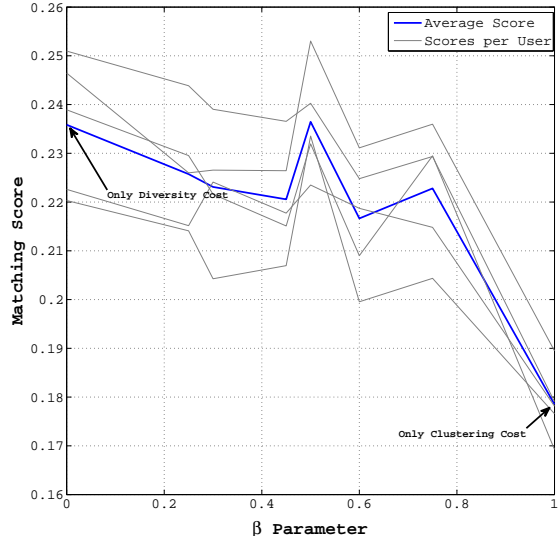}
\caption{\footnotesize{\textbf{Effect of the $\boldsymbol{\beta}$ parameter:} on summarization performance. It is interesting to note that when we make $\beta = 1$, there is a significant drop in the score since diversity is not considered at all. See algorithm \ref{algo:oprecis} for more details. Here results for $5$ different users at different $\beta$s are shown. The average is also depicted in bold.}}
\vspace{-10pt}
\label{fig:beta-effect}
\end{figure}

Algorithm \ref{algo:oprecis} describes the procedure to generate diverse samples in an online fashion. We initialize the exemplars with the first $K$ data points. Following this, we compute the diversity score for the current set of exemplars, denoted as \texttt{divscore}($\boldsymbol{\mu}$) in algorithm \ref{algo:oprecis}. Next, we begin the competition to find out which center has won the current round. Here winning is determined by a modified cost function that includes a diversity cost. The importance given to clustering error versus the diversity cost is governed by $\beta$, which is a user defined parameter. When $\beta = 1$, this expression reduces to the cost used in the online $K$-means algorithm. The effect of $\beta$ is shown in figure \ref{fig:beta-effect}, the right choice of $\beta$ can vary depending on the dataset and the features. Finally, we update the winning center only if it improves the overall diversity compared to the previous set. In some cases the centers may get stuck in local minimas, which can lead to poor exemplars. To avoid such cases, we add some noise, by updating centers even when they do not meet the diversity criterion in $1-10\%$ of the samples. 

\noindent \textbf{Complexity:} One of the main advantages of an online algorithm is that it can function with very low memory and computational resources. For the task of picking $K$ exemplars from dataset of $N$ points,  our algorithm requires $\mathcal{O}(K)$ for storage, compared to \emph{at least} $\mathcal{O}(N)$ for batch-mode summarization algorithms such as Precis \cite{ShroffTC10}. Typically, $N$ can be of the order of $10^5$ frames for an hour long video, whereas $K$ is typically around $10-50$. In terms of computational complexity, our algorithm takes $\mathcal{O}(NK)$ as compared to Precis \cite{ShroffTC10}, $\mathcal{O}(N(N-K)T)$ for $T$ iterations. When $N>>K$, which is typical in summarization, the computational complexity of our algorithm approximates to $\mathcal{O}(N)$ while Precis increases to $\mathcal{O}(N^2T)$. As a result, we are able to process features extracted from a video at about $14.3$ fps, in MATLAB on a standard Intel i7 PC. 
\vspace{-10pt}
\section{Experiments}
We perform experiments on the VSUMM dataset \cite{Avila2011}, which contains $50$ videos in MPEG-1 format (30 fps, 352 x 240 pixels), distributed across several genres (documentary, educational, ephemeral, historical, lecture) and their duration varies from 1 to 4 minutes and approximately 75 minutes of video in total \cite{VSUMMweb}. The dataset also contains $5$ different user evaluations per video, which are what human users have considered the best summary for the video. In order to exaggerate the advantage of using summarization over traditional sampling, we skew the dataset by replacing the last 500 frames of the video with a single \emph{frozen} frame. Such artifacts can be expected to occur, but more importantly they demonstrate the effectiveness of summarization. \\
\begin{figure*}[!htb]
\centering
\subfloat[\footnotesize{Diversity score for online K-medoids and the proposed algorithm, over 3 different videos. It is evident that our algorithm promotes diversity between exemplars much better than K-medoids.}]{
  \includegraphics*[clip = true,trim=45mm 10mm 52mm 15mm,width=0.37\linewidth]{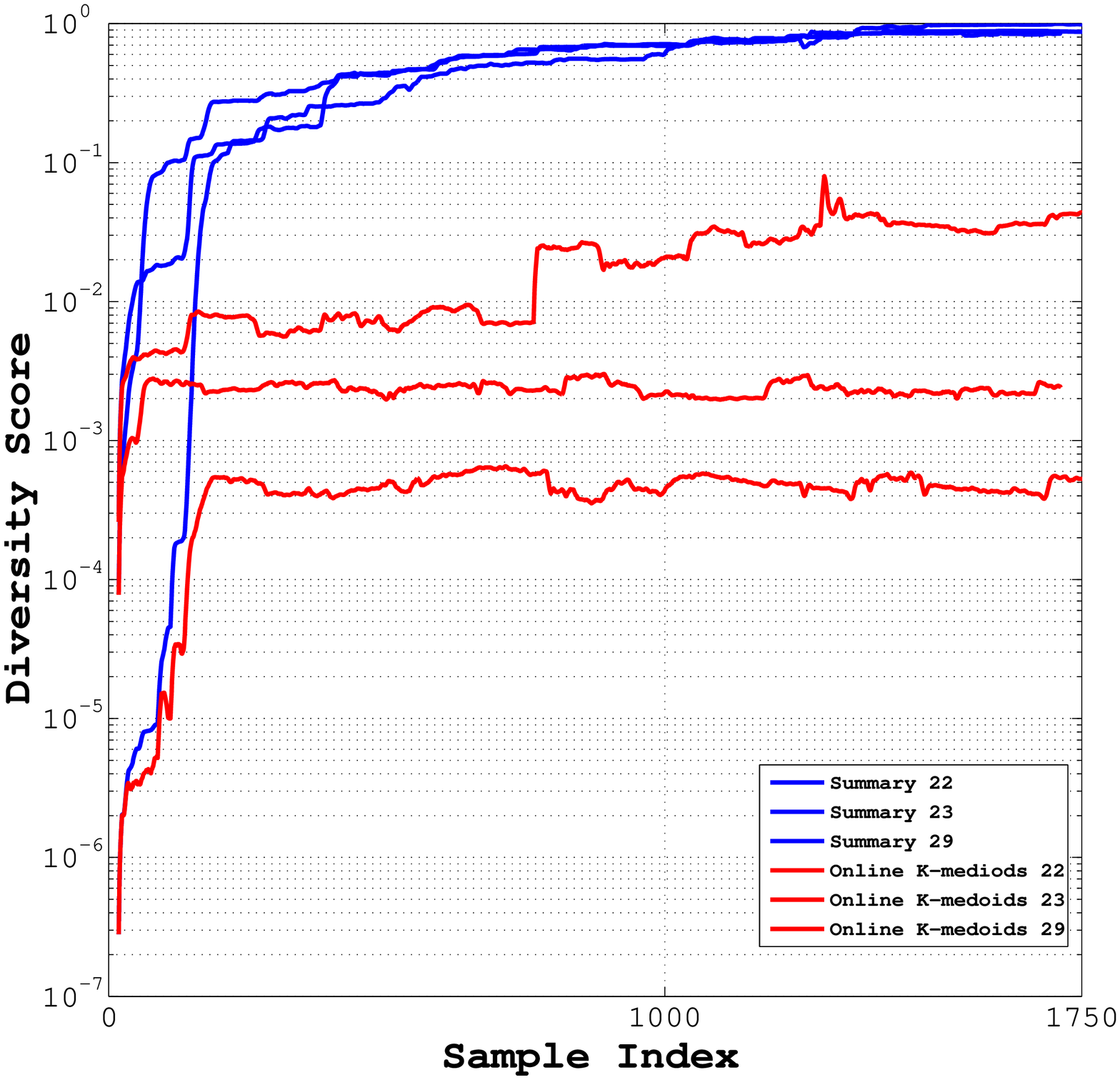}\label{fig:divcost}}
\subfloat[\footnotesize{Sample summaries generated for two different videos, the matches are marked in yellow.}]{
  \includegraphics*[clip = true,trim=30mm 40mm 30mm 35mm,width=0.63\linewidth]{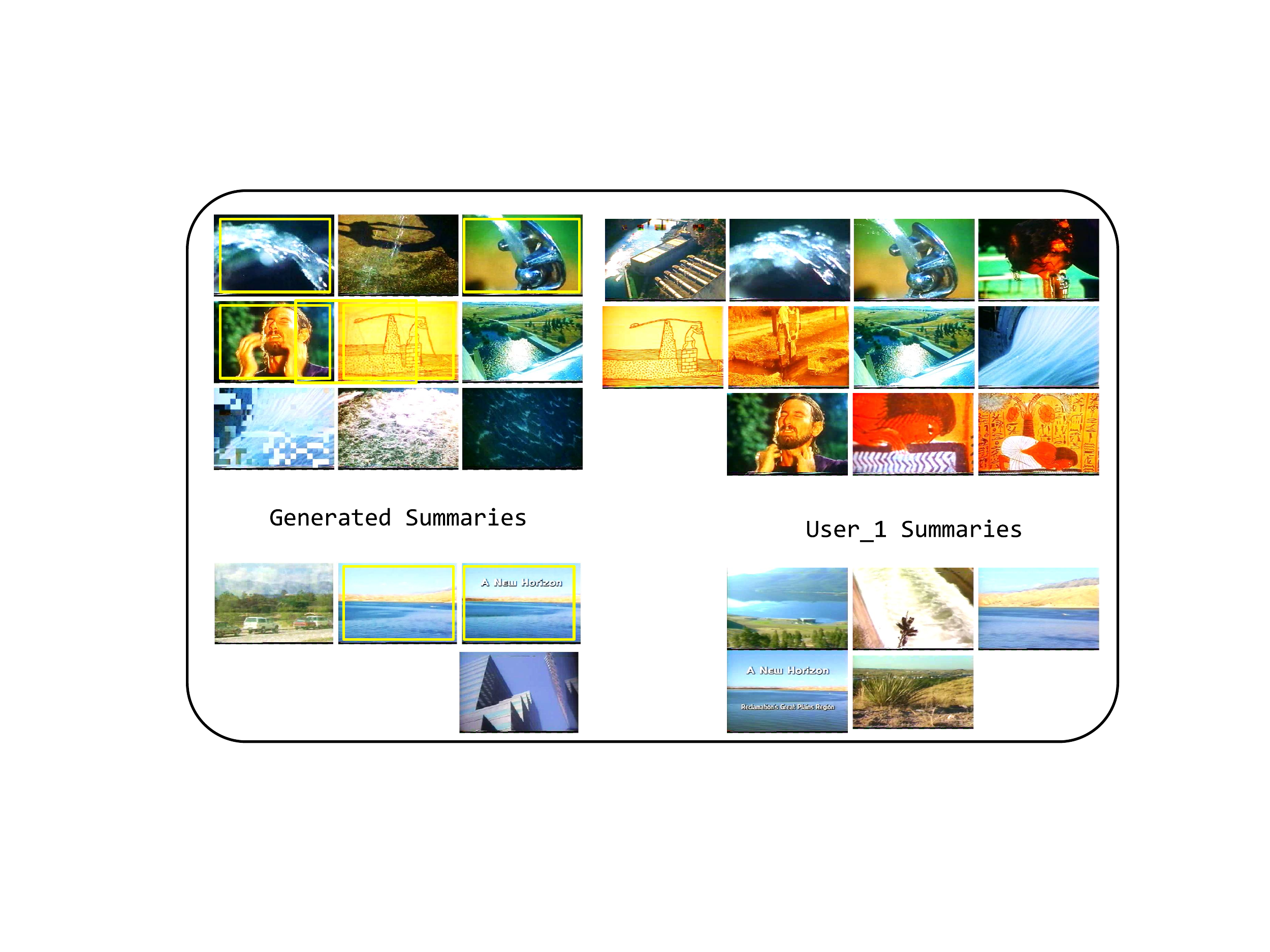}\label{fig:sample}}

\label{fig:examples}
\vspace{-15pt}
\end{figure*}
\noindent \textbf{Feature Extraction:} The video summarization problem is to pick the $K$ best exemplars from a set of $N$ points, $X = \{x_1,x_2,\dots,x_N\} \in \real^D$. The choice of $x_i$ is open to the application and the proposed algorithm can work with any kind of Euclidean features. We used deep features from the penultimate layer of a pre-trained neural network -- the VGG ``very deep'' network \cite{Simonyan2014vgg} trained on the ImageNet dataset \cite{ILSVRC15}. These pre-trained networks are available through the MatConvNet toolbox \cite{vedaldi15matconvnet}.  

\noindent {\bf Defining a match:} In order to accurately obtain the match score, we first filter the exemplars to remove similar frames. This is done by computing the $K\times K$ similarity matrix a set of exemplars, followed by picking only those points that have a distance greater than a fixed threshold, $\gamma$. The value for $\gamma$ needs to be chosen heuristically, and depends on the feature space. In our experiments with the deep features, we found that $\gamma = 70$, worked effectively in removing redundant exemplars. A weakness of using a fixed $\gamma$ is it may result in false positives and false negatives and better schemes maybe used to choose $\gamma$. To make a fair comparison, we use the same value of $\gamma$ across all our baselines.

\noindent \textbf{Evaluation:} Evaluating a summary is hard in general because there is no \emph{ground truth}. In many cases, the evaluation is done in comparison to human user generated summaries to find the highest ``matching'' score. In VSUMM \cite{Avila2011}, a new evaluation metric is proposed that measures the number of matching frames, and the number of non matching frames. The user generated summaries are of arbitrary lengths, as deemed suitable by the user. However, since our algorithm requires $K$, the number of desired exemplars as an input we modify the evaluation score to simply be the number of matches between each user generated summary and the summary generated by our algorithm. We choose $K$ to be equal the length of the largest summary set generated by a user per video, if $K<5$, then we set $K := 2*K$. This can be easily automated and chosen to be relative to the size of each video without affecting the results. Finally, we normalize the number of matches by the length of that user's summary.
\vspace{-10pt}
\subsection{Alternative Sampling Strategies and Results}
As a comparison to the proposed approach, we perform sampling using the following different baselines.\\
\noindent \textbf{Batch-mode Video Precis:} \cite{ShroffTC10} Our main comparison is with the Video Precis algorithm that optimizes between the representational error of the chosen samples and the diversity cost between any set of samples. The proposed algorithm can be considered an online version of Precis.

\noindent \textbf{Online K-medoids clustering:} We use the competitive learning algorithm used for online K-means (see \ref{sec:formulation}), as another comparison with comparable computational and memory complexity. Here, we set $\alpha = 1$, which is expected to be noisy since the learning rate is high. An alternative formulation could involve computing centers using a smaller $\alpha$, then assigning each center to the nearest data point. However, this violates the assumption of an online algorithm that does not have access to the entire dataset.

\noindent In addition we also report results using batch-mode K-medoids, random sampling and uniform sampling. Random and uniform sampling require knowledge of the number of frames or length of a video, which is unrealistic for streaming video. The performance of different sampling algorithms are reported in table \ref{tab:expt}, and it can be seen that the proposed diversity sampling performs better than batch mode summarization algorithm Precis. We are also significantly better than the online K-medoids algorithm and other baselines. Sample summaries are shown in figure \ref{fig:sample}, and the diversity score for our algorithm is compared to the diversity score obtained by the online K-medoids algorithm in figure \ref{fig:divcost}.

\begin{table}[!htb]
\centering  
\begin{tabular}{ |M{40pt}|M{17pt}|M{17pt}|M{17pt}|M{17pt}|M{17pt}|p{25pt}| } \hline
\small{Sampling Algorithm}& \small{U1} & \small{U2}& \small{U3} & \small{U4}&\small{U5}&\small{Online?}\\\hline
\small{K-medoids}	&0.191	&0.199&0.179&0.199 &0.193&\xmark \\\hline
Random 				&0.173 &0.165 &0.176 &0.186 & 0.179 &\xmark \\\hline
Uniform 			&0.190	& 0.196 & 0.188 & 0.200 & 0.193 &\xmark \\\hline
Precis \cite{ShroffTC10}&0.227 & 0.219 & 0.225  & 0.240 &\textbf{0.245}& \xmark  \\\hline\hline
\small{Online K-medoids}	& 0.141 & 0.129 & 0.131 & 0.146 &0.143&\cmark\\\hline
\textbf{Proposed}		&\textbf{0.240}& \textbf{0.224}& \textbf{0.234}& \textbf{0.253 }& 0.232&\cmark \\\hline
\end{tabular}
\vspace{-5pt}
 \caption{\footnotesize{Average mean scores denoting the percentage match with $5$ different users across 50 videos. The proposed online sampling scheme performs as well if not better than batch-mode Precis, and significantly outperforms comparable baselines.}}
 \label{tab:expt}
 \vspace{-15pt}
\end{table}

\section{Conclusion \& Future Work}
We presented the a novel online algorithm to perform streaming video summarization which can work with access to just a single frame at a time and does not need to know in advance the number of frames to allocate memory. We showed that the proposed online diverse sampling algorithm performs summarization as well as its batch-mode counter-parts, while being significantly more efficient. By generalizing aspects of competitive learning\cite{Desieno1988conscience}, and Video Precis \cite{ShroffTC10}, we are able to force the exemplars to be as \emph{diverse} as possible. We used PCA to map the centers to a lower dimensional space and then measured the volume of the convex hull in the PCA space as a measure of diversity. In the future, the dimensionality reduction step can be replaced with more advanced tools, that preserve topological properties and can potentially improve the robustness of the diversity measure. Another interesting extension is to generalize this algorithm to non Euclidean spaces such as Riemannian manifolds.

\bibliographystyle{IEEEbib}
\bibliography{ref}

\end{document}